\documentclass[11pt]{article}
\usepackage{booktabs} % For better table lines
\usepackage{makecell} % For more complex cell formats
\usepackage{eamt24}
\usepackage{times}
\usepackage{url}
\usepackage{latexsym}
\usepackage{todonotes}
\usepackage[small,bf]{caption} % added MLF 20171211
\title{Chasing COMET: Leveraging Minimum Bayes Risk Decoding for Self-Improving Machine Translation}

\author{Kamil Guttmann$\normalfont{\textsuperscript{* 1}}$, Mikołaj Pokrywka$\normalfont{\textsuperscript{* 1}}$, Adrian Charkiewicz$\normalfont{\textsuperscript{1}}$, Artur Nowakowski $\normalfont{\textsuperscript{1,2}}$ \\
        $\textsuperscript{1}$ Laniqo, Poznań, Poland \\
        $\textsuperscript{2}$ Faculty of Mathematics and Computer Science, Adam Mickiewicz University, Poznań, Poland \\
         {\tt \{name\}.\{surname\}@laniqo.com}}

\date{}

\begin{document}
\maketitle
\def\thefootnote{*}\footnotetext{Equal contribution}\def\thefootnote{\arabic{footnote}}

\begin{abstract}
This paper explores Minimum Bayes Risk (MBR) decoding for self-improvement in machine translation (MT), particularly for domain adaptation and low-resource languages. We implement the self-improvement process by fine-tuning the model on its MBR-decoded forward translations. By employing COMET as the MBR utility metric, we aim to achieve the reranking of translations that better aligns with human preferences. The paper explores the iterative application of this approach and the potential need for language-specific MBR utility metrics. The results demonstrate significant enhancements in translation quality for all examined language pairs, including successful application to domain-adapted models and generalisation to low-resource settings. This highlights the potential of COMET-guided MBR for efficient MT self-improvement in various scenarios.

\end{abstract}

\section{Introduction}
Machine translation (MT) bridges the gap between languages, fostering global communication and information exchange. However, achieving high-quality translations across diverse languages and domains remains a significant challenge, especially for low-resource languages where limited training data hinders model performance. Even in well-resourced settings, continuous improvement and adaptation to specific domains are ongoing research efforts.

This paper explores the potential of Minimum Bayes Risk (MBR) decoding~\cite{kumar-byrne-2004-minimum} as a self-improvement strategy for MT models. MBR decoding leverages the model's predictions to select the best translation from a set of candidates, potentially improving overall translation quality.

We employ COMET~\cite{rei-etal-2020-comet} as the utility function in MBR decoding and rerank candidate translations generated by an MT model. This approach creates a synthetic parallel dataset from monolingual data in the source language, enabling further model self-improvement.

This study examines the effectiveness of MBR decoding for self-improvement in three language pairs: English--German (high-resource), Czech--Ukrainian (low-resource), and English--Hausa (low-resource). For English--German, the focus is on the biomedical domain, incorporating additional monolingual data, while for Czech--Ukrainian, self-improvement is explored using only the training data translated by the model and reranked through MBR decoding. We further investigate the potential of iterative self-improvement with MBR decoding in both English--German and Czech--Ukrainian language pairs. Finally, in the case of English--Hausa, we compare the use of COMET, a massively multilingual metric, with a metric specifically tailored to African languages i.e. AfriCOMET~\cite{wang2023afrimte}. 

To determine the optimal configuration for MBR decoding, we investigate two decoding algorithms and various numbers of translation candidates.

\section{Related Work}

\paragraph{MBR and QE reranking with neural metrics}
MBR decoding, a technique commonly used in Statistical Machine Translation (SMT), has gained traction in Neural Machine Translation (NMT) in recent years.
\newcite{freitag-etal-2022-high} proposed using reference-based metrics, such as BLEURT ~\cite{sellam-etal-2020-bleurt} and Quality Estimation (QE) models, such as COMET-QE~\cite{rei-etal-2021-references} for reranking the set of hypotheses produced by the NMT model.

Similar work by~\newcite{fernandes-etal-2022-quality} proposed \textit{quality-aware decoding}. They explored various reranking strategies, including the well-performing pre-ranking of the set of hypotheses with QE models before passing them into MBR decoding. They found that using MERT-tuned~\cite{och-2003-minimum} reranker, where multiple QE metrics and model log-likelihood scores are linearly combined with learned weights to maximize a reference-based metric on a validation set shows improvements over the baseline.

\newcite{amrhein-sennrich-2022-identifying} used MBR decoding to identify biases and weaknesses in COMET, where they found that the early COMET models are not sufficiently sensitive to discrepancies in numbers and named entities.

MBR decoding performance is heavily dependent on the number of samples and the sampling strategy.
\newcite{freitag2023epsilon} investigated various sampling strategies and found that epsilon sampling outperformed others. This sampling method discards tokens with a probability below a certain threshold (epsilon), guaranteeing that each token in the final sample has a fair chance of being included. The approach is particularly effective when generating a large set of samples, as it inherently yields greater sample diversity compared to beam search.

\newcite{vernikos2024dont} introduced QE-fusion, a method that combines spans from different candidates sampled from a model using QE metrics. They found that the method consistently improves translation quality in terms of neural evaluation metrics, especially if applied to LLM due to their ability to generate diverse outputs.

Due to its ease of implementation and use, MBR and QE reranking have been successfully applied in machine translation shared tasks, as demonstrated by the results in several studies~\cite{nowakowski-etal-2022-adam,kudo-etal-2023-skim,jon-etal-2023-cuni}. This highlights its potential to significantly improve translation quality.

\paragraph{Model self-improvement}
Recent research has shown a growing interest in leveraging model outputs for self-improvement.
This approach holds significant promise in the case of machine translation, especially for low-resource and domain-specific translation scenarios, where there is access to the source-language data, but the corresponding target-language data is severely limited.

\newcite{gulcehre2023reinforced} describes reinforcement self-training (\textit{ReST}) method for language modeling. 
The method is based on producing a dataset for fine-tuning by sampling from the model (LLM). The samples are then scored with a QE metric. Then, offline reinforcement learning algorithms are applied using a reward-weighted loss based on the QE scores. The method can be applied to all generative learning settings, but the authors focus on its application to machine translation, showing that the method increases translation quality.

Concurrent work by~\newcite{finkelstein2023mbr} describes self-tuning NMT models on a set of hypotheses reranked using either MBR, QE, or a combination of the two methods.
They also experimented with using LLM as the teacher model, finding that it outperforms using a self-teacher and fine-tuning on references.

Our research expands on recent developments in the field by investigating the use of MBR-based fine-tuning in three key areas. Firstly, we examine its applicability in domain-specific translation tasks, specifically focusing on English--German translation in the biomedical domain. Secondly, we investigate its effectiveness for low-resource translation directions, exemplified by the Czech--Ukrainian language pair. This broadens the scope beyond English-centric language pairs, thus contributing to a more comprehensive analysis of MBR performance across less-represented languages in neural evaluation metrics. Finally, we explore the use of neural QE metrics tailored for specific languages, using AfriCOMET~\cite{wang2023afrimte} as an example.
\def\Statmt{Statmt}

\section{Experiment Overview}

\subsection{Model Self-Improvement}
The self-improvement process leverages MBR decoding to guide the model to select high-quality translations according to the utility function. The process consists of 3 steps:
\begin{description}
    \item[Step 1: Sample Generation] Using beam search decoding with beam size equal to \texttt{N}, generate \texttt{N} translation candidates using the base model for each source sentence. While \newcite{freitag2023epsilon} suggested that epsilon sampling might yield better results with MBR decoding, it typically requires reranking a significantly larger number of translation candidates, which becomes computationally expensive for processing large datasets. Beam search, on the other hand, allows for generating a smaller set of high-quality candidates while providing sufficient data for effective MBR decoding.
    \item[Step 2: MBR Decoding] Select a single translation for each source sentence from the list of candidates through MBR decoding utilizing COMET to guide the selection towards high-quality translations. For an efficient implementation of the MBR decoding algorithm, we use the code\footnote{\url{https://github.com/marian-nmt/marian-dev/tree/master/scripts/mbr}} from the Marian~\cite{mariannmt} framework.
    \item[Step 3: Model Fine-tuning] Fine-tune the base model on the synthetically created dataset. Use COMET as an early stopping metric during training to ensure fitting to this metric.
\end{description}

\subsection{English--German}
The English--German experiment simulates a real-world domain adaptation scenario. In such settings, while a large general-purpose parallel corpus might be available, the specific domain often lacks extensive parallel data. To address this challenge, we leveraged both a smaller parallel dataset and a larger monolingual dataset in the source language containing biomedical terminology.

To leverage the monolingual data in the source language we propose a two-step approach:
\begin{enumerate}
    \item Fine-Tuning: We fine-tune a general-purpose English--German model on a small parallel biomedical dataset.
    \item Self-improvement: To enhance the model performance in the biomedical domain, we incorporate a larger monolingual biomedical dataset during the self-improvement process. This involves creating a synthetic parallel dataset via MBR decoding and subsequently fine-tuning the biomedical translation model on the generated data.
\end{enumerate}

To assess the robustness of the self-improvement method, we conducted an additional experiment in which we applied this method to a model that was fine-tuned to the biomedical domain using general domain data for MBR decoding. This evaluated whether the model would retain its translation capabilities in the biomedical domain despite improvements based solely on out-of-domain data.

\subsection{Czech--Ukrainian}
The Czech--Ukrainian experiment addresses the challenge of machine translation between two low-resource languages. We aimed to evaluate whether self-improvement through MBR decoding leads to an increase in the overall translation quality when applied to language pairs that do not involve English, which typically dominate machine translation research.

In this setting, we used only the parallel data set without incorporating any additional monolingual data. To employ MBR decoding in this data-scarce environment, we directly translated the entire source side of the parallel dataset using the baseline translation model. This created a set of synthetic candidate translations, which were then reranked through MBR decoding.

In contrast to our English--German experiments where we incorporated external monolingual data, this setup explored self-improvement without relying on additional datasets. We achieved this by solely leveraging the information present within the data of the base model. This demonstrates the potential for self-improvement even in resource-constrained scenarios.

\subsection{English--Hausa}
The English--Hausa experiment delves into the critical question of how the choice of a quality evaluation metric influences the effectiveness of self-improvement with MBR decoding. We explored the impact of language coverage in the evaluation metric by comparing two approaches:
\begin{itemize}
    \item MBR decoding with WMT22 COMET: utilizing the \textit{wmt22-comet-da} model, which has been trained on direct assessments between a diverse set of language pairs.
    \item MBR decoding with AfriCOMET: using AfriCOMET-STL, a novel COMET-like metric specifically designed for evaluating translations to and from multiple African languages, including Hausa.
\end{itemize}

The objective of this study was to investigate the effect of language contribution in the neural evaluation metric on the quality of translations decoded using MBR. The comparison of these two approaches specifically addresses whether self-improvement guided by the WMT22 COMET metric, which is trained on a diverse range of language pairs, can effectively generalize to low-resource language pairs. Furthermore, we explore the potential need to use language-specific metrics, such as AfriCOMET-STL for Hausa, to achieve better performance in such scenarios.

\subsection{Iterative MBR Self-Improvement}

Following the initial self-improvement through MBR decoding, we explored the possibility of applying it iteratively to further enhance the model's translation quality.

We started each iteration by selecting the best model checkpoint based on the WMT22 COMET metric on the validation set. Next, we performed MBR decoding on the entire training set using this checkpoint, generating a new iteration of the synthetic training set. Finally, we resumed the training of the model using the new training set, starting from the previously selected checkpoint.

The iterative process was repeated until a decrease was observed in the evaluation scores of metrics other than WMT22 COMET. In the case of English--German biomedical translation, the process was continued until the model's quality improved solely on an in-domain test set and decreased on a general domain test set, as this could indicate potential overfitting to the biomedical domain.

\section{Experimental Setup}

\subsection{Data Filtering}
We filtered the general training data using the following heuristic filters:
\begin{itemize} \label{filtering}
  \item average length of words in each sentence (character-wise) $\leq 15$;
  \item number of characters in each sentence $\leq 500$;
  \item digits in a sentence (character-wise)  $\leq 15\%$;
  \item number of characters in the longest word $\leq 28$;
  \item number of words in sentence $\leq 100$;
  \item Levenshtein distance between source and target sentences $\geq 2$;
  \item number of characters in each sentence $\geq 5$;
  \item probability that each sentence is in the correct language $\geq 10\%$.
\end{itemize}

To ensure that each sentence is in the correct language, we have used the fastText LID-201 language identification model~\cite{burchell-etal-2023-open}.

The Bicleaner-AI model~\cite{zaragoza-bernabeu-etal-2022-bicleaner} is also used to filter the English--German dataset. This tool estimates the likelihood that a sentence pair constitutes a mutual translation. A threshold of 50\% is established for the Bicleaner score within this language pair. Bicleaner-AI is not utilized for other language pairs due to the unavailability of open-source models for those languages.

% ---- VOCAB ----
\subsection{Vocabulary}
We employed SentencePiece~\cite{kudo-richardson-2018-sentencepiece}, a subword tokenization library, to train unigram tokenizers for each language pair in our experiments.

For the English--German and English--Hausa setups, we created a joint vocabulary containing 32,000 subword tokens and tied all embeddings during the training of the MT model. In contrast, for Czech--Ukrainian, due to different scripts (Latin and Cyrillic), we created separate vocabularies of 32,000 subword tokens and tied only the target and output layer embeddings.

% ---- BASELINE MODEL PARAMETERS ----
\subsection{Baseline Model Hyperparameters}
For all experiments, we trained Transformer (big)~\cite{NIPS2017_3f5ee243} models using the Marian framework. These models were trained on four NVIDIA A100 GPUs, each equipped with 80GB of VRAM.

Hyperparameter Settings:
\begin{itemize} \label{hyperparameters}
    \item learning rate: 2e-4;
    \item learning rate warmup: 8000 updates;
    \item learning rate decay: inverse square root;
    \item mini-batch size determined automatically to fit GPU memory;
    \item early stopping after 10 consecutive validations with no improvement in mean word cross-entropy score.
\end{itemize}

\subsection{Evaluation metrics}

We use sacreBLEU~\cite{post-2018-call} to calculate BLEU\footnote{\scriptsize BLEU signature: nrefs:1$\vert$case:mixed$\vert$eff:no$\vert$tok:13a$\vert$smooth:exp$\vert$version:2.3.1}~\cite{papineni-etal-2002-bleu} and chrF\footnote{\scriptsize chrF signature: nrefs:1$\vert$case:mixed$\vert$eff:yes$\vert$nc:6$\vert$nw:0$\vert$space:no$\vert$version:2.3.1}~\cite{popovic-2015-chrf}. 

We acknowledge the potential for overfitting to the WMT22 COMET\footnote{\url{https://huggingface.co/Unbabel/wmt22-comet-da}} metric used for MBR decoding. Therefore, we extended the evaluation to also include CometKiwi\footnote{\url{https://huggingface.co/Unbabel/wmt22-cometkiwi-da}}~\cite{rei-etal-2022-cometkiwi}, UniTE\footnote{\url{https://huggingface.co/Unbabel/unite-mup}}~\cite{wan-etal-2022-unite}, UniTE-DA\footnote{\url{https://huggingface.co/Unbabel/wmt22-unite-da}}~\cite{rei-etal-2023-inside} and BLEURT-20\footnote{\url{https://storage.googleapis.com/bleurt-oss-21/BLEURT-20.zip}}~\cite{sellam2020bleurt}. 

For the English--Hausa experiments, we additionally calculated scores using AfriCOMET-STL \cite{wang2023afrimte}, which was specifically trained to evaluate translations involving certain African languages.

\subsection{English to German}
\def\krsTest{khresmoi}
\def\floresTest{FLORES-200}
To train the baseline model, we used all corpora from the MTData toolkit (version 0.4.0) \cite{gowda-etal-2021-many}, excluding the validation sets and the test sets from the available datasets. Our filters described in Section~\ref{filtering} reduced the dataset from approximately 800 million sentences to 400 million. \label{general-data}

In the context of domain adaptation, we employed the following list of domain data:
\begin{enumerate}
    \item 40 thousand sentences from biomedical-translation-corpora~\cite{neves-etal-2016-scielo};
    \item 3 million sentences from Ufal medical corpus shared in WMT23~\cite{kocmi-etal-2023-findings};
    \item 2 million sentences from EMEA corpus downloaded from OPUS~\cite{tiedemann-nygaard-2004-opus}.
\end{enumerate}
After deduplication, we were left with 3 million sentences which we split into two datasets. We considered a scenario with 1 million bilingual parallel sentences and approximately 2 million monolingual sentences in the source language. Khresmoi-dev~\cite{krs-testset} concatenated with \floresTest{}~\cite{nllb-22} was utilized as the validation set during training. We did not apply any filtering to the domain data.
    
\def\baseGen{\textbf{Baseline}}
\def\mixTune{\textbf{Mix-tune}}
\def\baseMbr{\textbf{Base-domain-mbr}}
\def\mixTuneMbr{\textbf{Mix-tune-domain-mbr}}
\def\mixTuneMbrIter2{\textbf{Mix-tune-domain-mbr-iter2}}
\def\mixTuneGenMbr{\textbf{Mix-tune-general-mbr}}

\def\baseGenb{Baseline}
\def\mixTuneb{Mix-tune}
\def\baseMbrb{Base-domain-mbr}
\def\mixTuneMbrb{Mix-tune-domain-mbr}
\def\mixTuneMbrIterb{Mix-tune-domain-mbr-iter2}
\def\mixTuneGenMbrb{Mix-tune-general-mbr}

We used the above data to train the following models:
\begin{itemize}
    \item Baseline (\baseGen) -- model trained only on data from the MTdata toolkit.
    \item Baseline + mix-tuning (\mixTune) -- fine-tuned \baseGen{} model on 1 million in-domain bilingual data concatenated with 1 million general-domain data randomly sampled from the \baseGen{} training set.
    \item Baseline + domain MBR (\baseMbr) -- fine-tuned \baseGen{} model on 2 million domain-specific sentences from MBR-decoded forward translations.
    \item Mix-tuned + domain MBR (\mixTuneMbr) -- fine-tuned \mixTune{} model on 2 million domain-specific sentences from MBR-decoded forward translations.
    \item Mix-tuned + MBR-iteration2 (\textbf{\mixTuneMbrIter2}) -- fine-tuned \mixTuneMbr{} on the 2 million domain-specific sentences from MBR-decoded forward translations.
    \item Mix tuned + general-MBR (\mixTuneGenMbr) -- fine-tuned \mixTune{} model on 2 million sentences sampled from the general-domain corpora from the \baseGen{} training set as MBR-decoded forward translations.
\end{itemize}
When fine-tuning the \mixTune{} model, we tailor the learning rate setup to meet specific requirements: learn-rate: 1e-7, lr-decay-inv-sqrt: 16000, lr-warmup: 16000. All remaining fine-tuning procedures employ an adjusted learning rate set to \mbox{5e-6}. 

\subsection{Czech to Ukrainian}

We leveraged all of the Czech--Ukrainian parallel data from the WMT23 MTData recipe, resulting in approximately 8 million sentence pairs after filtering as described in Section~\ref{filtering}. We did not include any additional monolingual data in this experiment.

We utilized the FLORES-200 dataset for validation during training, while the WMT22 test set served as an additional benchmark.

We trained the baseline model only on the parallel data, using hyperparameters as described in Section~\ref{hyperparameters}. Next, we translated the source side of the parallel corpus used in training with our baseline model, saving a list of translation candidates. We performed MBR decoding, selecting the best translation of each set of candidate translations, resulting in a synthetic training dataset.

We investigated the following approaches to leverage the MBR-decoded data for model improvement:
\begin{itemize} \label{mbr-approaches}
    \item Standard fine-tuning (\textbf{MBR-finetuned}) -- we fine-tuned the baseline model on the MBR-decoded data, using a learning rate of ~5e-6.
    \item Fine-tuning with a high learning rate (\textbf{MBR-ft-high-lr}) -- we fine-tune the baseline model on MBR-decoded data, using a learning rate of 2e-4.
    \item Resuming training with MBR-decoded data (\textbf{MBR-resumed}) -- we switched the training set to the MBR-decoded version and resumed training, restoring the optimizer state and effectively continuing its training with the improved data.
\end{itemize}

\subsection{English to Hausa}
To train the models in the English--Hausa direction, we used data from the WMT shared tasks from previous years. Specifically, we used:
\begin{enumerate}
    \item 7 million sentences from OPUS;
    \item 2.4 million data from the WMT23 African MT Shared Task ~\cite{kocmi-etal-2023-findings};
    \item 150 thousand sentences from ParaCrawl v8.0~\cite{banon-etal-2020-paracrawl}.
\end{enumerate}
The deduplication process reduced the data size to approximately 9 million sentences. Following the filtering criteria detailed in Section~\ref{filtering}, a total of 3.1 million sentences were retained. We used FLORES-200 for validation during training. After training, we evaluated the model on the FLORES-200 and NTREX test sets.

We took similar steps as in the Czech--Ukrainian experiment, training a baseline model with hyperparameters set as described in Section~\ref{hyperparameters}. We conducted experiments employing MBR decoding, comparing its performance using two distinct metrics as the utility function:
\begin{itemize}
    \item WMT22 COMET -- based on XLM-RoBERTa~\cite{conneau-etal-2020-unsupervised}, covering a diverse set of 100 languages,
    \item AfriCOMET-STL -- based on AfroXLM-RoBERTa~\cite{alabi-etal-2022-adapting}, covering 17 African languages and 3 high-resource languages.
\end{itemize}

We investigated the impact of the chosen metric for MBR decoding by training two models using the refined translations:
\begin{itemize}
    \item \textbf{MBR-COMET} -- training resumed with the training set switched to the WMT22 COMET MBR-decoded version.
    \item \textbf{MBR-AfriCOMET} -- training resumed with the training set switched to the AfriCOMET-STL MBR-decoded version.
\end{itemize}

\section{Results}
The statistical significance of the evaluation results is assessed using a paired bootstrap resampling test~\cite{koehn-2004-statistical}, involving 1000 resampling trials to confirm the statistical significance of the model improvements (\textit{p} $< 0.05$).
\subsection{Number of translation samples and search algorithm}

To determine the optimal setup for MBR decoding, we conducted experiments involving the translation and evaluation of chosen test sets with various MBR decoding sample sizes and two decoding algorithms. This approach offers the advantages of being both representative and computationally efficient compared to training MT models on the entire MBR-decoded training set.

We evaluated two decoding algorithms -- beam search and top-k. For the top-k setup, we experimented with temperature values of 0.1 and 1, keeping the \textit{k} parameter equal to 10. These choices were based on the work 
 done by~\newcite{freitag2023epsilon}. To determine the best number of samples for MBR decoding we conducted experiments with the following numbers of samples: 10, 25, 50, 100, 200, 300, 400, 500.

Firstly we noted that beam search is the preferred option, given its high scores and greater stability across different metric results, as observed in Figure~\ref{fig:en_de_krs_topk_bs} and~\ref{fig:cs_uk_flores_topk_bs}. We provide more specific results in the Appendix Figures~\ref{fig:krs-bs-topk},~\ref{fig:flores-bs-topk}.

\begin{figure}[hbt!]
 \centering
 \includegraphics[scale=0.35]{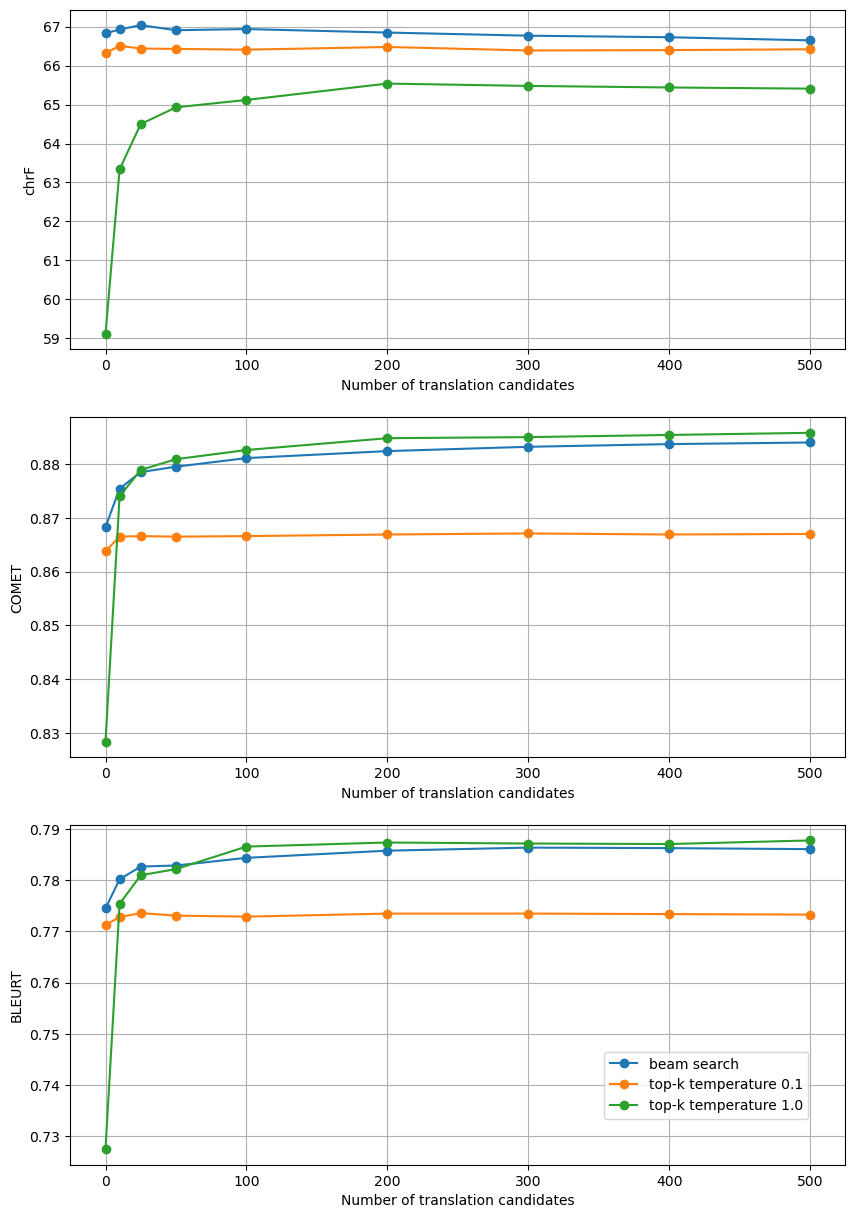}
 \caption{Comparison of beam search and top-k algorithms of the \mixTune{} English--German model for the \krsTest{} test set. Top-k algorithm with temperature 1.0 showed superior performance on neural metrics over top-k with temperature 0.1 and slightly better performance than beam search. However, beam search achieved the highest score on the chrF metric, while the top-k algorithm with temperature 1.0 had the lowest score (translation without MBR decoding is represented on the chart as the number of translation candidates equal to 0).}\label{fig:en_de_krs_topk_bs}
\end{figure}

\begin{figure}[hbt!]
 \centering
 \includegraphics[scale=0.35]{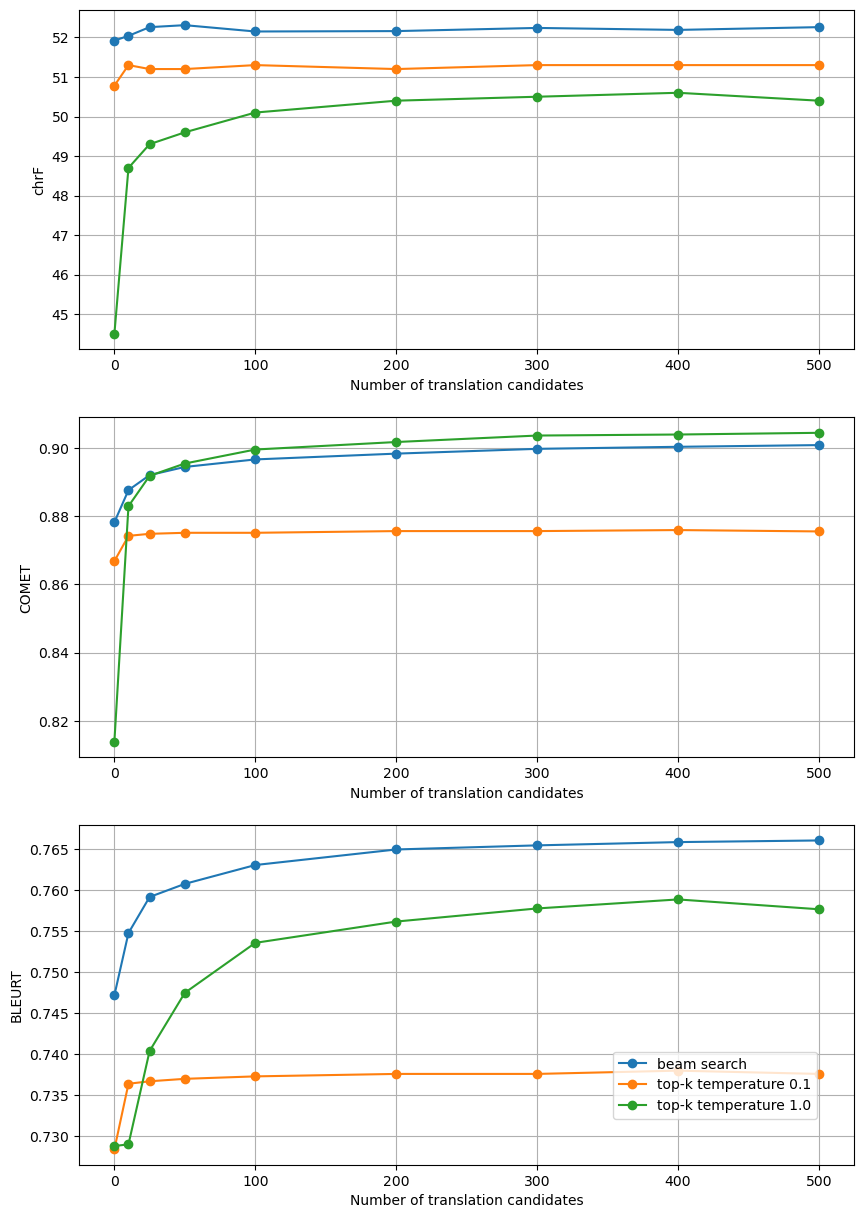}
 \caption{Comparison of beam search and top-k algorithms of the \textbf{baseline} Czech--Ukrainian model for the \mbox{\floresTest{}} test set. Beam search seems to be the superior option with the best performance on chrF and BLEURT metrics and slightly worse results on COMET over top-k with temperature 1.0 (translation without MBR decoding is represented on the chart as the number of translation candidates equal to 0).}\label{fig:cs_uk_flores_topk_bs}
\end{figure}

Secondly, we decided to train our models on MBR-decoded data from 50 candidates selected by the beam search decoding algorithm. We considered the balance between improvement in evaluation metrics based on neural language models, stability across lexical metrics, and the execution time of MBR decoding, as shown in Figure~\ref{fig:en_de_krs_topk_bs_all_metrics}. \vfill \null  We provide more detailed results in the Appendix Figures~\ref{fig:krs-bs},~\ref{fig:flores-en-de-bs},~\ref{fig:krs-topk01},~\ref{fig:krs-topk10},~\ref{fig:cs-uk-bs},~\ref{fig:cs-uk-topk01},~\ref{fig:cs-uk-topk10}.

\begin{figure*}[hbt!]
 \centering
 \includegraphics[scale=0.5]{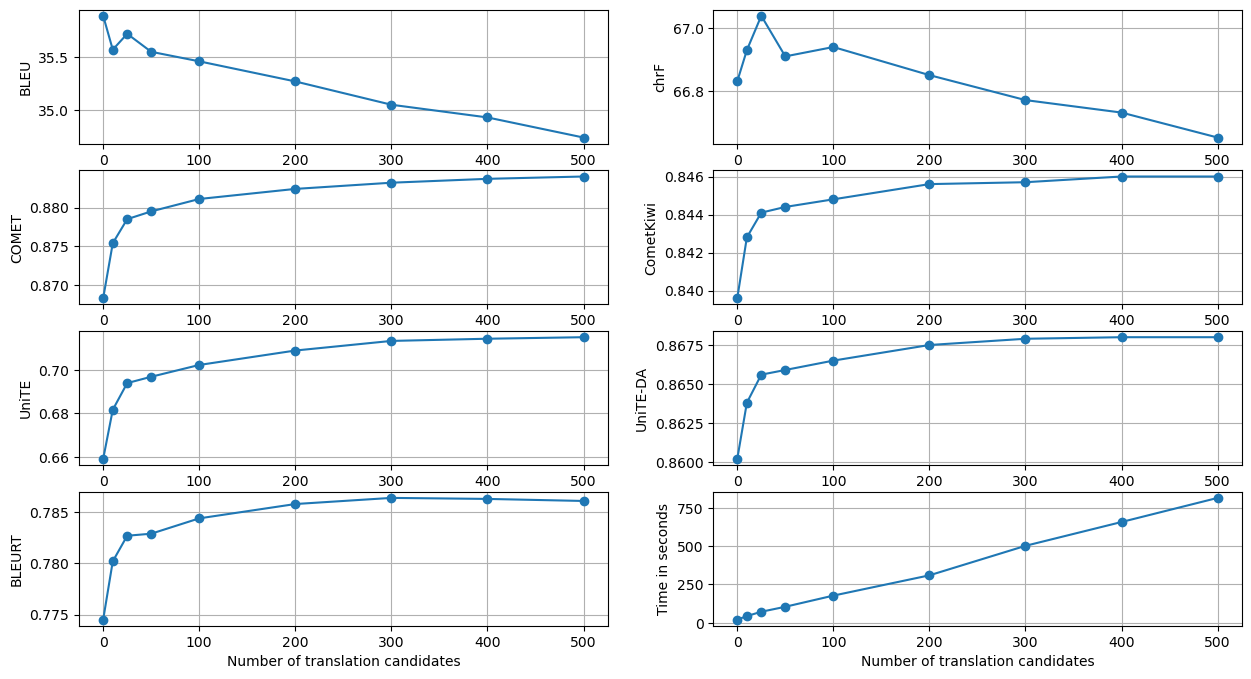}
 \caption{Comparison of beam search performance with a different number of samples of the \mixTune{} English--German model for the \krsTest{} test set. Initial increases in the number of samples for MBR decoding showed very rapid gains, but further increases no longer resulted in such large gains, and performance on the n-gram metrics deteriorated (translation without MBR decoding is represented on the chart as the number of translation candidates equal to 0).}\label{fig:en_de_krs_topk_bs_all_metrics}
\end{figure*}

\subsection{English to German}
Table \ref{tab:en-de-krs} shows the evaluation results on the in-domain test set \krsTest{}. All models self-improved with MBR decoding have shown enhanced performance. However, model \mixTuneMbrIter2{} did not exhibit improvement over its first iteration \mixTuneMbr{}, even on COMET, which was the utility metric of MBR decoding. \mixTuneGenMbr{} model shows a slightly better performance on BLEURT metric compared to models fine-tuned on in-domain MBR-decoded forward translations. 

\begin{table}[hbt!]
\centering
\scalebox{0.8}{
\begin{tabular}{lccc}
\toprule
Model                          & chrF & COMET  & BLEURT \\ 
\midrule
\baseGenb{}         & 66.6          & 0.8653          & 0.7693          \\
\mixTuneb{}         & 66.8          & 0.8682          & 0.7749          \\
\baseMbrb{}         & \textbf{66.9} & 0.8711*          & 0.7755          \\
\mixTuneMbrb{}      & \textbf{66.9} & \textbf{0.8728}* & 0.7792*          \\
\mixTuneMbrIterb{} & \textbf{66.9} & 0.8727*          & 0.7791*          \\
\mixTuneGenMbrb{}   & \textbf{66.9} & 0.8720*          & \textbf{0.7799}* \\
\bottomrule
\end{tabular}}
\caption{English--German \krsTest{} set results for the MBR self-improvement approaches. All models fine-tuned with MBR self-improvement technique have shown better performance over \baseGen{} and \mixTune{} models, including the \mixTuneGenMbr{} model, which was finetuned on general-domain MBR-decoded data. The results marked with an asterisk (*) are statistically significant compared to the \mixTune{} model.} 
\label{tab:en-de-krs}
\end{table}

Table \ref{tab:en-de-flores} presents the evaluation results on the FLORES-200 test set. Although chrF did not increase, the neural evaluation metrics showed improvement. Similar to the \krsTest{} test set, the \mixTuneMbrIter2 model showed a decrease in quality during the second iteration of self-improvement. \mixTuneGenMbr{} showed superior performance over other models.

\begin{table}[hbt!]
\centering
\scalebox{0.8}{
\begin{tabular}{lccc}
\toprule
Model              & chrF & COMET  & BLEURT \\
\midrule
\baseGenb{}         & \textbf{67.5} & 0.8751          & 0.7735          \\
\mixTuneb{}         & \textbf{67.5} & 0.8756          & 0.7744          \\
\baseMbrb{}         & 67.2          & 0.8772          & 0.7743          \\
\mixTuneMbrb{}      & 67.3          & 0.8787*          & 0.7766          \\
\mixTuneMbrIterb{} & 67.1          & 0.8766          & 0.7748          \\
\mixTuneGenMbrb{}   & \textbf{67.5} & \textbf{0.8813}* & \textbf{0.7784}* \\
\bottomrule
\end{tabular}}
\caption{English--German \floresTest{} test set results for the MBR self-improvement approaches. \mixTuneGenMbr{} model has shown superior performance, however, models with domain-specific forward translation maintain performance. The results marked with an asterisk (*) are statistically significant compared to the \mixTune{} model.}
\label{tab:en-de-flores}
\end{table}

\def\medline{WMT22-medline}

In summary, our findings demonstrate that applying MBR decoding significantly improves the performance of the high-resource English--German model for low-resource biomedical domain translation, particularly on neural network metrics. While lexical metrics show lower stability, they also hold potential for improvement.

Experiments demonstrated the robustness of self-improving models with the MBR decoding technique. Model fine-tuned on general forward translation had great performance on the in-domain test set and the model fine-tuned on domain-specific forward translation maintained performance on the general domain test set. We provide a broader evaluation in the Appendix Tables~\ref{tab:en-de-krs-spec},~\ref{tab:en-de-wmt22medline-spec},~\ref{tab:en-de-flores-spec},~\ref{tab:en-de-statmt-spec}.

\subsection{Czech to Ukrainian}

\begin{table}[hbt!]
\centering
\begin{tabular}{lccc}
\toprule
Model & chrF & COMET & BLEURT \\
\midrule
Baseline & 52.0 & 0.8779 & 0.7466 \\ 
MBR-finetuned & 52.4 & 0.8839 & 0.7522 \\
MBR-ft-high-lr & \textbf{52.7} & \textbf{0.8869} & 0.7553 \\
MBR-resumed & \textbf{52.7} & 0.8864 & \textbf{0.7557} \\
\bottomrule
\end{tabular}
\caption{Czech--Ukrainian FLORES-200 test set results for the three MBR self-improvement approaches. All self-improved models exhibit improvements on all metrics compared to the baseline model, regardless of the fine-tuning approach used. Notably, both \textbf{MBR-ft-high-lr} and \textbf{MBR-resumed} models achieve the highest gains, demonstrating comparable performance. All self-improved models show statistical significance compared to the \textbf{Baseline} model.}
\label{tab:cs-uk-flores}
\end{table}

\begin{table}[hbt!]
\centering
\begin{tabular}{lccc}
\toprule
Model & chrF & COMET & BLEURT \\
\midrule
Baseline & 58.4 & 0.8721 & 0.7498 \\
MBR-finetuned & 60.0 & 0.8803 & 0.7574 \\
MBR-ft-high-lr & \textbf{60.2} & 0.8844 & 0.7619 \\
MBR-resumed & 60.0 & \textbf{0.8852} & \textbf{0.7639} \\
\bottomrule
\end{tabular}
\caption{Czech--Ukrainian WMT22 test set results for the three MBR self-improvement approaches. Similar to the FLORES-200 results, all self-improved models exhibit improvements on all metrics compared to the baseline model. However, on the WMT22 test set, the neural metrics favour the \textbf{MBR-resumed} model over the \textbf{MBR-ft-high-lr} model. All self-improved models show statistical significance compared to the \textbf{Baseline} model.}
\label{tab:cs-uk-wmt22}
\end{table}

\begin{table}[hbt!]
\centering
\setlength{\tabcolsep}{4pt}
\begin{tabular}{lccc}
\toprule
Model & chrF  & COMET & BLEURT \\
\midrule
Baseline & 52.0 & 0.8779 & 0.7466 \\ 
MBR-resumed & 52.7* & 0.8864* & 0.7557* \\
MBR-resumed-iter2 & \textbf{52.8} & 0.8888* & \textbf{0.7567} \\
MBR-resumed-iter3 & 52.6 & \textbf{0.8901} & 0.7557 \\     
\bottomrule
\end{tabular}
\caption{Czech--Ukrainian iterative self-improvement results on the FLORES-200 test set. While the COMET score consistently improves across all three iterations, the chrF and BLEURT scores show a decrease in the third iteration. This suggests that the model overfits to COMET, harming the quality of the translation. Results with an asterisk (*) are statistically significant in comparison with the model in the row directly above it.}
\label{tab:cs-uk-flores-iterative}
\end{table}

\begin{table}[hbt!]
\centering
\setlength{\tabcolsep}{4pt}
\begin{tabular}{lccc}
\toprule
Model & chrF & COMET & BLEURT \\
\midrule
Baseline & 58.4 & 0.8721 & 0.7498 \\
MBR-resumed & 60.0* & 0.8852* & 0.7639* \\
MBR-resumed-iter2 & \textbf{60.3}* & 0.8885* & \textbf{0.7641} \\
MBR-resumed-iter3 & 60.1 & \textbf{0.8896} & 0.7578 \\
\bottomrule
\end{tabular}
\caption{Czech--Ukrainian iterative self-improvement results on the WMT22 test set. Consistent with the FLORES-200 results, the COMET score improves across all iterations, while other metrics show a decrease in the last iteration. Notably, the BLEURT score not only decreases but falls below the score achieved by the first self-improved model. Results with an asterisk (*) are statistically significant in comparison with the model in the row directly above it.}
\label{tab:cs-uk-wmt-iterative}
\end{table}

The results of the three MBR self-improvement approaches described in Section \ref{mbr-approaches} are presented in Tables \ref{tab:cs-uk-flores} and \ref{tab:cs-uk-wmt22} for the FLORES-200 and WMT22 test sets, respectively.

We find that standard fine-tuning of the baseline model with MBR-decoded data yields the smallest improvements across all metrics, suggesting its limited effectiveness in this context. We note that both fine-tuning with a higher learning rate and resuming the training exhibit comparable performance, with resumed training achieving slightly better results on the WMT22 test set. This may indicate that resuming training helps mitigate overfitting to the FLORES-200 validation set used during training.

Tables \ref{tab:cs-uk-flores-iterative} and \ref{tab:cs-uk-wmt-iterative} showcase the impact of iterative training with MBR decoding on the FLORES-200 and WMT22 test sets, respectively. The second iteration consistently improves scores across all metrics, demonstrating the effectiveness of the iterative self-improvement process in refining the model's translation capabilities. However, the third iteration leads to a decrease in both chrF and BLEURT scores. This suggests potential overfitting to the MBR decoding utility metric, where the model prioritizes aspects that score well according to COMET but may not translate to overall translation quality.

We provide extended evaluations in the Appendix in Tables \ref{tab:cs-uk-flores-appendix}, \ref{tab:cs-uk-wmt-appendix}, \ref{tab:cs-uk-flores-iterative-appendix}, \ref{tab:cs-uk-wmt-iterative-appendix}.

\subsection{English to Hausa}

\begin{table}[hbt!]
\centering
\setlength{\tabcolsep}{4pt}
\scalebox{0.8}{
\begin{tabular}{@{}lcccc@{}}
\toprule
Model & chrF & COMET & BLEURT & AfriCOMET \\
\midrule
Baseline & 49.9 & 0.7569 & 0.7931 & 0.6984 \\
MBR-COMET & 50.9 & \textbf{0.7720} & \textbf{0.8083} & 0.7207 \\
MBR-AfriCOMET & \textbf{51.2} & 0.7692 & 0.8061 & \textbf{0.7239} \\
\bottomrule
\end{tabular}
}
\caption{English--Hausa FLORES-200 test set results for MBR self-improvement with different metrics. Both self-improved models achieve gains compared to the baseline model on all evaluation metrics. While the AfriCOMET-based model achieves a higher AfriCOMET score, reflecting its alignment with the specific evaluation metric, the COMET-based model surpasses it in both BLEURT and COMET scores, while showing a comparable gain on the AfriCOMET score. All self-improved models show statistical significance compared to the \textbf{Baseline} model.}
\label{tab:en-ha-flores}
\end{table}

\begin{table}[hbt!]
\centering
\setlength{\tabcolsep}{4pt}
\scalebox{0.8}{
\begin{tabular}{@{}lcccc@{}}
\toprule
Model & chrF & COMET & BLEURT & AfriCOMET  \\
\midrule
Baseline & 51.6 & 0.7596 & 0.7791 & 0.6800 \\
MBR-COMET & \textbf{53.1} & \textbf{0.7752} & \textbf{0.7986} &  0.7046 \\
MBR-AfriCOMET & 53.0 & 0.7721 & 0.7956 & \textbf{0.7062} \\
\bottomrule
\end{tabular}
}
\caption{English--Hausa NTREX test set results for MBR self-improvement with different metrics. Similar to the FLORES-200 results, both self-improved models using MBR decoding demonstrate improvements over the baseline model on all evaluation metrics. All self-improved models show statistical significance compared to the \textbf{Baseline} model.}
\label{tab:en-ha-ntrex}
\end{table}

This section compares the performance of two MBR decoding self-improvement approaches for English--Hausa translation: one utilizing the WMT22 COMET model and another using the AfriCOMET model. The results are presented in Tables \ref{tab:en-ha-flores} and \ref{tab:en-ha-ntrex} for the FLORES-200 and NTREX test sets, respectively.

We observe that the AfriCOMET MBR-tuned model achieves gains over the WMT22 COMET MBR-tuned model on chrF for the FLORES-200 test set, but this advantage is not replicated on the NTREX test set. Additionally, the gains from AfriCOMET MBR-tuning are mainly limited to the AfriCOMET metric.

Our analysis reveals that the \textbf{MBR-AfriCOMET} model exhibits improvements over the \textbf{MBR-COMET} model primarily on lexical metrics in the case of the FLORES-200 test set, but not in the case of NTREX. The gains of the \textbf{MBR-AfriCOMET} model are mainly limited to AfriCOMET metrics, while other neural-based metrics consistently favour the \textbf{MBR-COMET} model.

While WMT22 COMET might exhibit a lower correlation with human judgment for the English--Hausa language pair than AfriCOMET, as reported by~\newcite{wang2023afrimte}, both self-improved models achieved significant and comparable gains on AfriCOMET. This suggests that WMT22 COMET, can still correctly rerank translation candidates and effectively guide the self-improvement process, leading to improvements on AfriCOMET, a metric specifically designed for African languages. This finding suggests that self-improvement guided by WMT22 COMET, with its diverse language coverage, might be effective even in low-resource settings, potentially reducing the need for additional adaptation of neural evaluation models to individual languages.

Additional evaluations are provided in the Appendix in Tables \ref{tab:en-ha-flores-appendix}, \ref{tab:en-ha-ntrex-appendix}.

\section{Conclusion}

This study demonstrated the effectiveness of model self-improvement through MBR decoding in improving translation quality. This approach proves beneficial for both high and low-resource languages, offering versatility in its application across diverse scenarios. Examples include domain-specific translation and the enhancement of general translation models.

We conducted experiments with various sample sizes for MBR decoding, using two decoding algorithms: beam search and top-k. The aim was to find a balance between automatic metric gains and time efficiency. Our experiments have shown that the beam search algorithm with a beam size set to 50 is the optimal choice.

In the field of high-resource English-to-German biomedical translation, we investigated the impact of domain adaptation using various self-improvement approaches on MBR-decoded forward-translated data. Experiments showed that all MBR-based fine-tuning, regardless of the domain of the test set, improved performance compared to the baseline model. This finding highlights the robustness of the self-improvement technique.

Experiments on the Czech--Ukrainian language pair revealed that fine-tuning the MT model on MBR-decoded translations of the training data set significantly improves translation performance. Applying this process iteratively improves quality, but further iterations yield diminishing gains and at some point, the quality may even degrade due to overfitting to the MBR decoding utility metric.

In the English--Hausa experiments, we employed two models for MBR decoding: WMT22 COMET and AfriCOMET. Both models yielded comparable and significant improvements in automatic metrics, indicating their effectiveness in guiding the self-improvement process. While AfriCOMET, specifically trained on African languages, might intuitively seem favourable for this language pair, the performance of the \textbf{MBR-COMET} model highlights the potential of utilizing more widely applicable metrics like WMT22 COMET even for low-resource settings.

% \bibliography{\confname}

\bibliography{eamt24}
\bibliographystyle{eamt24}

\appendix

\begin{table*}[hbt!]
\centering
\scalebox{0.95}{
\begin{tabular}{lccccccc}
\toprule
Model              & chrF          & BLEU          & COMET           & CometKiwi       & UniTE        & UniTE-DA        & BLEURT          \\
\midrule
\baseGenb{}         & 66.6          & 35.6          & 0.8653          & 0.8373          & 0.6441          & 0.8574          & 0.7693          \\
\mixTuneb{}         & 66.8         & \textbf{35.9} & 0.8682          & 0.8397          & 0.6594          & 0.8602          & 0.7749          \\
\baseMbrb{}         & \textbf{66.9} & 35.7          & 0.8711          & 0.8416          & 0.6694          & 0.8621          & 0.7755          \\
\mixTuneMbrb{}{}      & \textbf{66.9} & 35.8          & \textbf{0.8728} & \textbf{0.8423} & 0.6766 & 0.8631          & 0.7792          \\
\mixTuneMbrIterb{} & \textbf{66.9} & 35.6          & 0.8727          & \textbf{0.8423} & 0.6757          & \textbf{0.8633} & 0.7791          \\
\mixTuneGenMbrb{}   & \textbf{66.9} & 35.5          & 0.8720          & 0.8422          & \textbf{0.6775}          & 0.8631          & \textbf{0.7799} \\
\bottomrule
\end{tabular}
}
\caption{English--German \krsTest{} set results for the MBR self-improvement approaches. All models fine-tuned with MBR self-improvement technique have shown better performance over \baseGen{} and \mixTune{} models, even \mixTuneGenMbr{} model with general forward translations.}
\label{tab:en-de-krs-spec}
\end{table*}

\begin{table*}[hbt!]
\centering
\scalebox{0.95}{
\begin{tabular}{lccccccc}
\toprule
Model              & chrF          & BLEU          & COMET           & CometKiwi       & UniTE        & UniTE-DA        & BLEURT          \\
\midrule
\baseGenb{}         & 63.1          & 35.0          & 0.8505          & 0.8336          & 0.5368          & 0.8470          & 0.7500          \\
\mixTuneb{}         & 63.5          & 35.6          & 0.8525          & 0.8360          & 0.5418          & 0.8495          & 0.7541          \\
\baseMbrb{}         & 63.5          & 35.8          & \textbf{0.8549}          & 0.8374          & 0.5549          & 0.8501          & 0.7522          \\
\mixTuneMbrb{}      & 63.6          & 35.7          & 0.8540          & 0.8379          & 0.5552          & 0.8508          & 0.7530          \\
\mixTuneMbrIterb{} & \textbf{63.7} & \textbf{35.9} & 0.8543          & \textbf{0.8383} & 0.5575          & \textbf{0.8510} & 0.7535          \\
\mixTuneGenMbrb{}   & 63.4          & 35.4          & 0.8547 & 0.8378          & \textbf{0.5613} & 0.8501          & \textbf{0.7542} \\
\bottomrule
\end{tabular}
}
\caption{English--German \medline{} set results for the MBR self-improvement approaches. All models fine-tuned with MBR self-improvement technique have shown better performance over \mixTune{} model except on metric BLEURT. On this specific test set, \mixTuneMbrIter2{} outperformed the \mixTuneMbr{} model, unlike the results observed on other test sets.}
\label{tab:en-de-wmt22medline-spec}
\end{table*}

\begin{table*}[hbt!]
\centering
\scalebox{0.95}{
\begin{tabular}{lccccccc}
\toprule
Model              & chrF          & BLEU          & COMET           & CometKiwi       & UniTE        & UniTE-DA        & BLEURT          \\
\midrule
\baseGenb{}         & \textbf{67.5} & 42.0          & 0.8751          & 0.8454          & 0.6630          & 0.8614          & 0.7735          \\
\mixTuneb{}         & \textbf{67.5} & \textbf{42.2} & 0.8756          & 0.8457          & 0.6657          & 0.8617          & 0.7744          \\
\baseMbrb{}         & 67.2          & 41.7          & 0.8772          & 0.8469          & 0.6677          & 0.8632          & 0.7743          \\
\mixTuneMbrb{}      & 67.3          & 41.7          & 0.8787          & 0.8477          & 0.6719          & 0.8641          & 0.7766          \\
\mixTuneMbrIterb{} & 67.1          & 41.5          & 0.8766          & 0.8466          & 0.6653          & 0.8629          & 0.7748          \\
\mixTuneGenMbrb{}   & \textbf{67.5} & 41.8          & \textbf{0.8813} & \textbf{0.8484} & \textbf{0.6824} & \textbf{0.8654} & \textbf{0.7784} \\
\bottomrule
\end{tabular}
}
\caption{English--German \floresTest{} test set results for the MBR self-improvement approaches. \mixTuneGenMbr{} model has shown superior performance, however, models with domain-specific forward translation maintain performance.}
\label{tab:en-de-flores-spec}
\end{table*}

\begin{table*}[hbt!]
\centering
\scalebox{0.95}{
\begin{tabular}{lccccccc}
\toprule
Model              & chrF          & BLEU          & COMET           & CometKiwi       & UniTE        & UniTE-DA        & BLEURT          \\
\midrule
\baseGenb{}         & 63.8          & 36.6          & 0.8428          & 0.8328          & 0.5308          & 0.8420          & 0.7106          \\
\mixTuneb{}         & 63.7          & 36.5          & 0.8427          & 0.8322          & 0.5283          & 0.8414          & 0.7107          \\
\baseMbrb{}         & 63.3          & 35.8          & 0.8463          & 0.8359          & 0.5376          & 0.8454          & 0.7138          \\
\mixTuneMbrb{}      & 63.2          & 35.9          & 0.8468          & 0.8358          & 0.5404          & 0.8464          & 0.7132          \\
\mixTuneMbrIterb{} & 63.0          & 35.5          & 0.8460          & 0.8345          & 0.5348          & 0.8455          & 0.7119          \\
\mixTuneGenMbrb{}   & \textbf{64.1} & \textbf{36.7} & \textbf{0.8629} & \textbf{0.8399} & \textbf{0.5622} & \textbf{0.8492} & \textbf{0.7202} \\
\bottomrule
\end{tabular}
}
\caption{English--German \Statmt{} test set results for the MBR self-improvement approaches. \mixTuneGenMbr{} model has shown significantly improved performance on every metric, however models with domain-specific forward translation maintain performance.}
\label{tab:en-de-statmt-spec}
\end{table*}

\begin{table*}[hbt!]
\centering
\begin{tabular}{lccccccc}
\toprule
Model & chrF & BLEU & COMET & CometKiwi & UniTE & UniTE-DA & BLEURT \\
\midrule
Baseline & 52.0 & 22.2 & 0.8779 & 0.8449 & 0.4441 & 0.9017 & 0.7466 \\ 
MBR-finetuned & 52.4 & 22.3 & 0.8839 & 0.8513 & 0.4715 & 0.9063 & 0.7522 \\
MBR-ft-high-lr & \textbf{52.7} & 22.6 & \textbf{0.8869} & \textbf{0.8543} & \textbf{0.4829} & 0.9085 & 0.7553 \\
MBR-resumed & \textbf{52.7} & \textbf{22.8} & 0.8864 & 0.8540 & 0.4824 & \textbf{0.9086} & \textbf{0.7557} \\
\bottomrule
\end{tabular}
\caption{Extended Czech--Ukrainian FLORES-200 test set results for the three MBR self-improvement approaches. All approaches lead to an increase in evaluation scores. Both \textbf{MBR-ft-high-lr} and \textbf{MBR-resumed} models achieve the highest gains, demonstrating comparable performance. }
\label{tab:cs-uk-flores-appendix}
\end{table*}

\begin{table*}[hbt!]
\centering
\begin{tabular}{lccccccc}
\toprule
Model & chrF & BLEU & COMET & CometKiwi & UniTE & UniTE-DA & BLEURT \\
\midrule
Baseline & 58.4 & 31.1 & 0.8721 & 0.8046 & 0.3744 & 0.8795 & 0.7498 \\
MBR-finetuned & 60.0 & 32.3 & 0.8803 & 0.8121 & 0.4112 & 0.8846 & 0.7574 \\
MBR-ft-high-lr & \textbf{60.2} & \textbf{33.2} & 0.8844 & 0.8152 & \textbf{0.4246} & 0.8880 & 0.7619 \\
MBR-resumed & 60.0 & 33.0 & \textbf{0.8852} & \textbf{0.8162} & 0.4236 & \textbf{0.8890} & \textbf{0.7639} \\
\bottomrule
\end{tabular}
\caption{Extended Czech--Ukrainian WMT22 test set results for the three MBR self-improvement approaches. As in the case of evaluation results on the FLORES-200 test set, all approaches improve upon the baseline model, although \textbf{MBR-resumed} stands out across all neural metrics apart from UniTE.}
\label{tab:cs-uk-wmt-appendix}
\end{table*}

\begin{table*}[hbt!]
\centering
\begin{tabular}{lccccccc}
\toprule
Model & chrF  & BLEU & COMET & CometKiwi & UniTE & UniTE-DA & BLEURT \\
\midrule
Baseline & 52.0 & 22.2 & 0.8779 & 0.8449 & 0.4441 & 0.9017 & 0.7466 \\ 
MBR-resumed & 52.7 & \textbf{22.8} & 0.8864 & 0.8540 & 0.4824 & 0.9086 & 0.7557 \\
MBR-resumed-iter2 & \textbf{52.8} & 22.6 & 0.8888 & 0.8557 & \textbf{0.4882} & \textbf{0.9099} & \textbf{0.7567} \\
MBR-resumed-iter3 & 52.6 & 22.3 & \textbf{0.8901} & \textbf{0.8562} & 0.4873 & 0.9097 & 0.7557 \\     
\bottomrule
\end{tabular}
\caption{Extended Czech--Ukrainian iterative self-improvement results on the FLORES-200 test set. Models increase in quality across all neural metrics until the third iteration, when the quality measured by metrics other than COMET and CometKiwi decreases. It's worth noticing that the BLEU score increases only in the first iteration and slowly degrades in consecutive iterations. }
\label{tab:cs-uk-flores-iterative-appendix}
\end{table*}

\begin{table*}[hbt!]
\centering
\begin{tabular}{lccccccc}
\toprule
Model & chrF & BLEU & COMET & CometKiwi & UniTE & UniTE-DA & BLEURT \\
\midrule
Baseline & 58.4 & 31.1 & 0.8721 & 0.8046 & 0.3744 & 0.8795 & 0.7498 \\
MBR-resumed & 60.0 & \textbf{33.0} & 0.8852 & 0.8162 & 0.4236 & 0.8890 & 0.7639 \\
MBR-resumed-iter2 & \textbf{60.3} & 32.6 & 0.8885 & \textbf{0.8183} & \textbf{0.4349} & \textbf{0.8900} & \textbf{0.7641} \\
MBR-resumed-iter3 & 60.1 & 31.9 & \textbf{0.8896} & 0.8174 & 0.4312 & 0.8887 & 0.7578 \\
\bottomrule
\end{tabular}
\caption{Extended Czech--Ukrainian iterative self-improvement results on the WMT22 test set. Evaluations across all metrics show similar tendencies as in the case of FLORES-200, except for CometKiwi which also decreases in the third iteration.}
\label{tab:cs-uk-wmt-iterative-appendix}
\end{table*}

\begin{table*}[hbt!]
\centering
\setlength{\tabcolsep}{3pt}
\begin{tabular}{lcccccccc}
\toprule
Model & chrF & BLEU & COMET & CometKiwi & UniTE & UniTE-DA & BLEURT & AfriCOMET \\
\midrule
Baseline & 49.9 & 22.3 & 0.7569 & 0.5597 & -0.2297 & 0.6082 & 0.7931 & 0.6984 \\
MBR-COMET & 50.9 & 23.2 & \textbf{0.7720} & \textbf{0.5707} & \textbf{-0.1777} & \textbf{0.6233} & \textbf{0.8083} & 0.7207 \\
MBR-AfriCOMET & \textbf{51.2} & \textbf{23.4} & 0.7692 & 0.5638 & -0.1878 & 0.6183 & 0.8061 & \textbf{0.7239} \\
\bottomrule
\end{tabular}
\caption{Extended English--Hausa results on the FLORES-200 test set. According to lexical metrics and AfriCOMET, the \textbf{MBR-AfriCOMET} model shows the greatest improvement. However, other neural metrics suggest that the \textbf{MBR-COMET} model is superior.}
\label{tab:en-ha-flores-appendix}
\end{table*}

\begin{table*}[hbt!]
\centering
\setlength{\tabcolsep}{3pt}
\begin{tabular}{lcccccccc}
\toprule
Model & chrF & BLEU & COMET & CometKiwi & UniTE & UniTE-DA & BLEURT & AfriCOMET \\
\midrule
Baseline & 51.6 & 23.9 & 0.7596 & 0.5704 & -0.1763 & 0.6294 & 0.7791 & 0.6800 \\
MBR-COMET & \textbf{53.1} & \textbf{25.3} & \textbf{0.7752} & \textbf{0.5865} & \textbf{-0.1051} & \textbf{0.6484} & \textbf{0.7986} & 0.7046 \\
MBR-AfriCOMET & 53.0 & 24.9 & 0.7721 & 0.5803 & -0.1273 & 0.6409 & 0.7956 & \textbf{0.7062} \\
\bottomrule
\end{tabular}
\caption{Extended English--Hausa results on the NTREX test set. In contrast to evaluations on the FLORES-200 test set, in this case only the AfriCOMET metric favours the \textbf{MBR-AfriCOMET} model.}
\label{tab:en-ha-ntrex-appendix}
\end{table*}

\begin{figure*}[hbt!]
 \centering
 \includegraphics[scale=0.5]{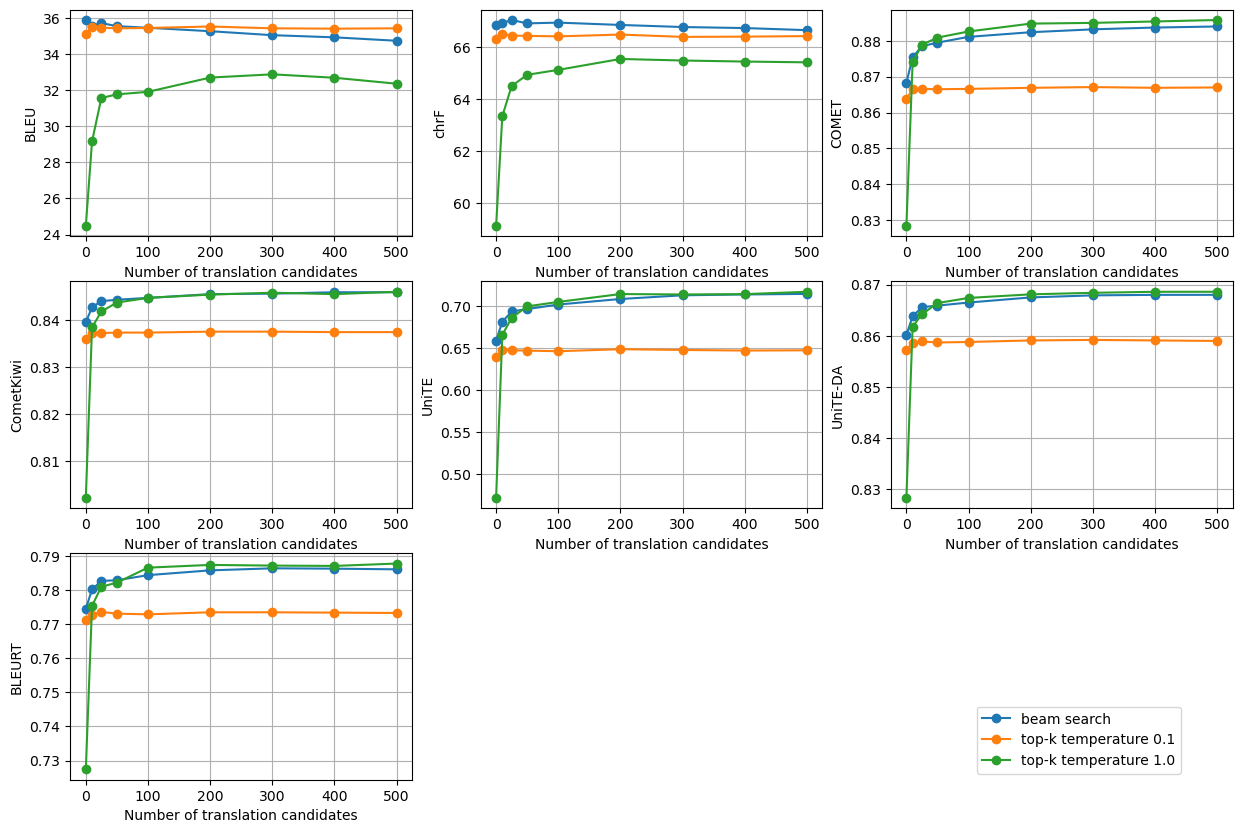}
 \caption{Comparison of beam search and top-k algorithms of the \mixTune{} English--German model for the \krsTest{} test set. Top-k algorithm with temperature 1.0 showed superior performance on neural metrics over top-k with temperature 0.1 and slightly better performance than beam search. However, beam search achieved the highest score on the chrF metric, while the top-k algorithm with temperature 1.0 had the lowest score for lexical metrics (translation without MBR decoding is represented on the chart as the number of translation candidates equal to 0).}\label{fig:krs-bs-topk}
\end{figure*}

\begin{figure*}[hbt!]
 \centering
 \includegraphics[scale=0.5]{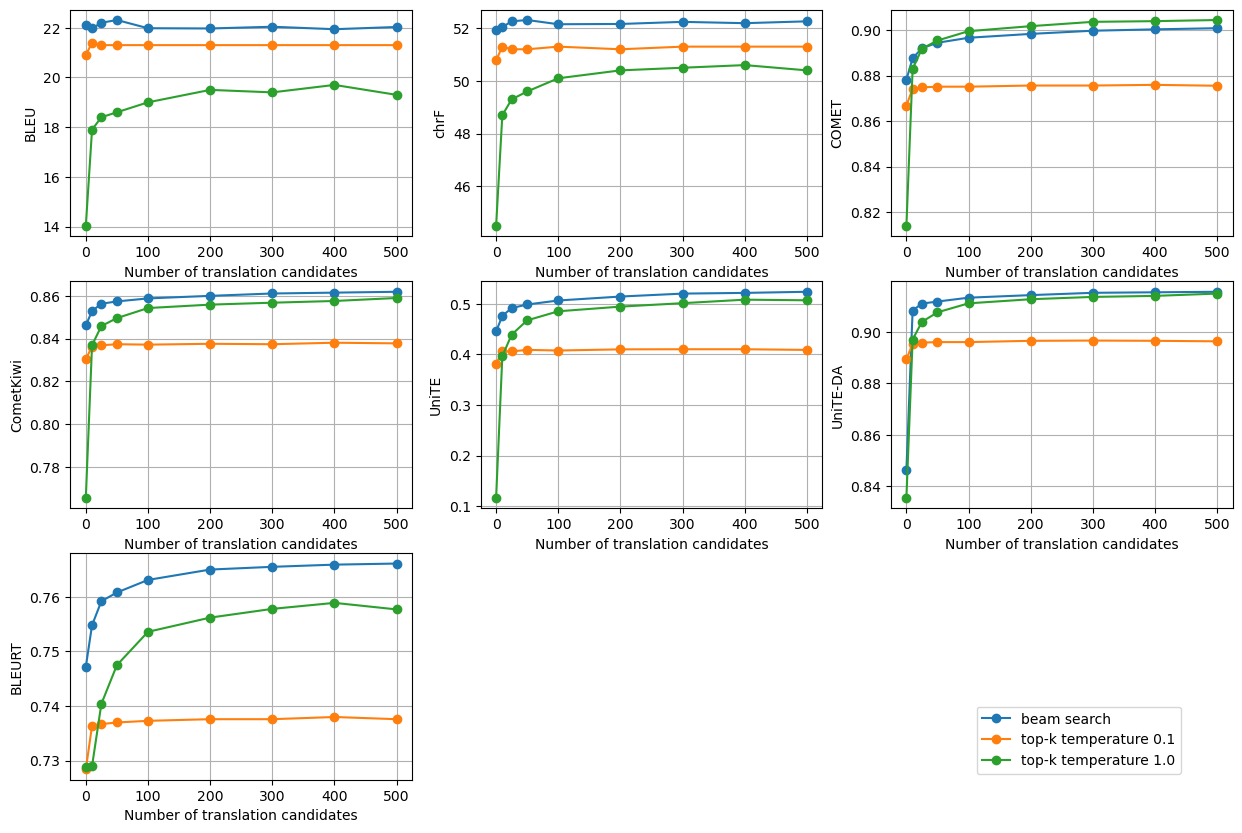}
 \caption{Comparison of beam search and top-k algorithms of the \textbf{baseline} Czech--Ukrainian model for the \floresTest{} test set. Beam search seems to be the superior option with the best performance on every metric except COMET (translation without MBR decoding is represented on the chart as the number of translation candidates equal to 0).}\label{fig:flores-bs-topk}
\end{figure*}

\begin{figure*}[hbt!]
 \centering
 \includegraphics[scale=0.5]{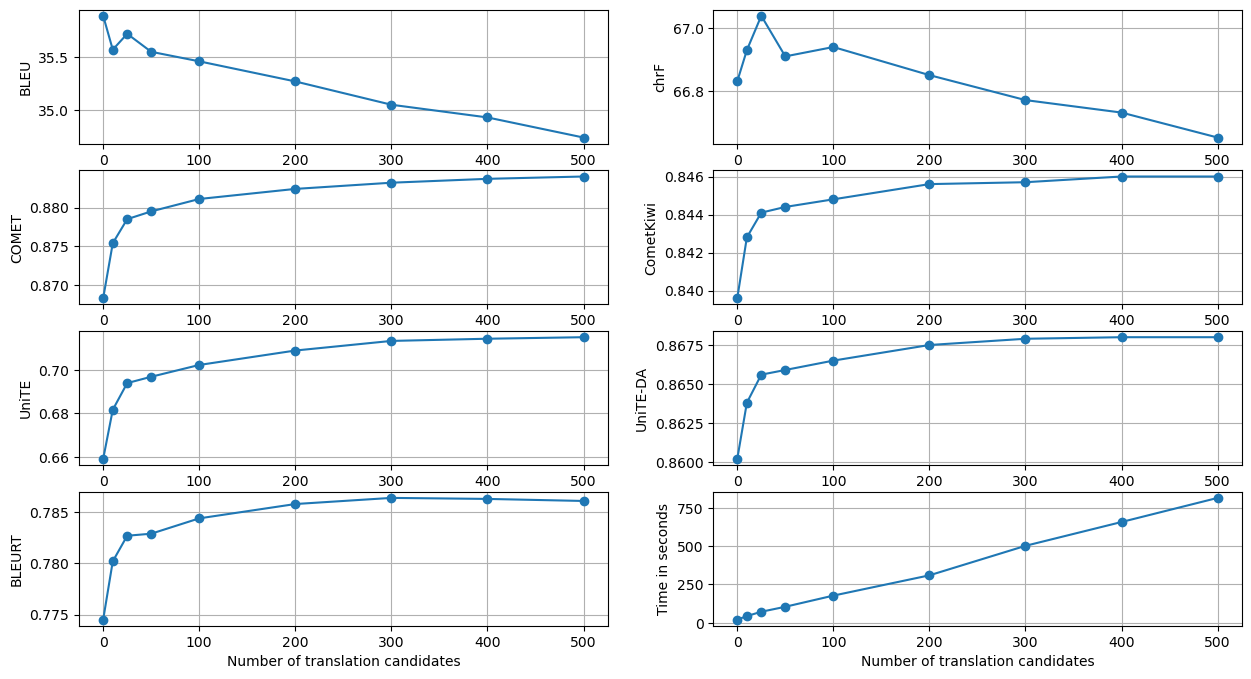}
 \caption{Comparison of beam search performance with a different number of samples of the \mixTune{} English--German model for the \krsTest{} test set. Initial increases in the number of samples for MBR decoding showed very rapid gains, but further increases no longer resulted in such large gains and performance on the n-gram metrics deteriorated (translation without MBR decoding is represented on the chart as the number of translation candidates equal to 0).}\label{fig:krs-bs}
\end{figure*}

\begin{figure*}[hbt!]
 \centering
 \includegraphics[scale=0.5]{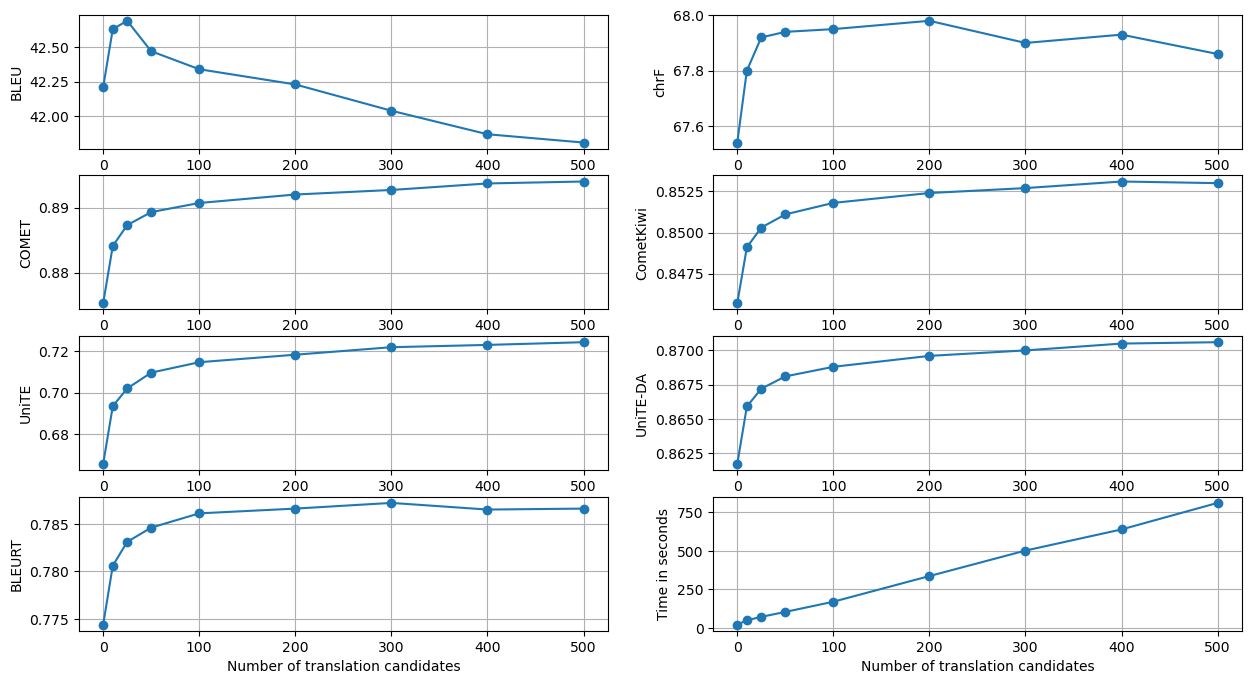}
 \caption{Comparison of beam search performance with a different number of samples of the \mixTune{} English--German model for the \floresTest{} test set. Initial increases in the number of samples for MBR decoding showed very rapid gains, but further increases no longer resulted in such large gains and performance on the n-gram metrics deteriorated (translation without MBR decoding is represented on the chart as the number of translation candidates equal to 0).}\label{fig:flores-en-de-bs}
\end{figure*}

\begin{figure*}[hbt!]
 \centering
 \includegraphics[scale=0.5]{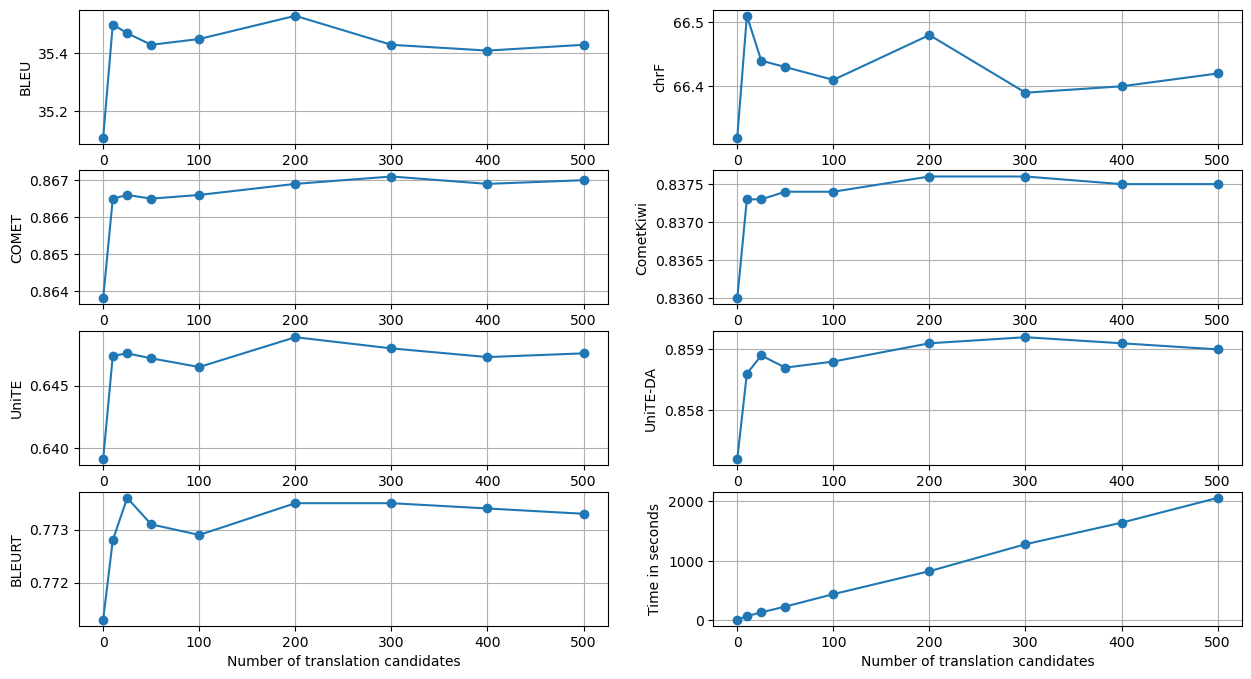}
 \caption{Comparison of top-k performance (temperature 0.1, k=10) with different number of samples of the \mixTune{} English--German model for the \krsTest{} test set. Initial increases in the number of samples for MBR decoding showed very rapid gains, but further increases no longer resulted in such large gains (translation without MBR decoding is represented on the chart as the number of translation candidates equal to 0).}\label{fig:krs-topk01}
\end{figure*}

\begin{figure*}[hbt!]
 \centering
 \includegraphics[scale=0.5]{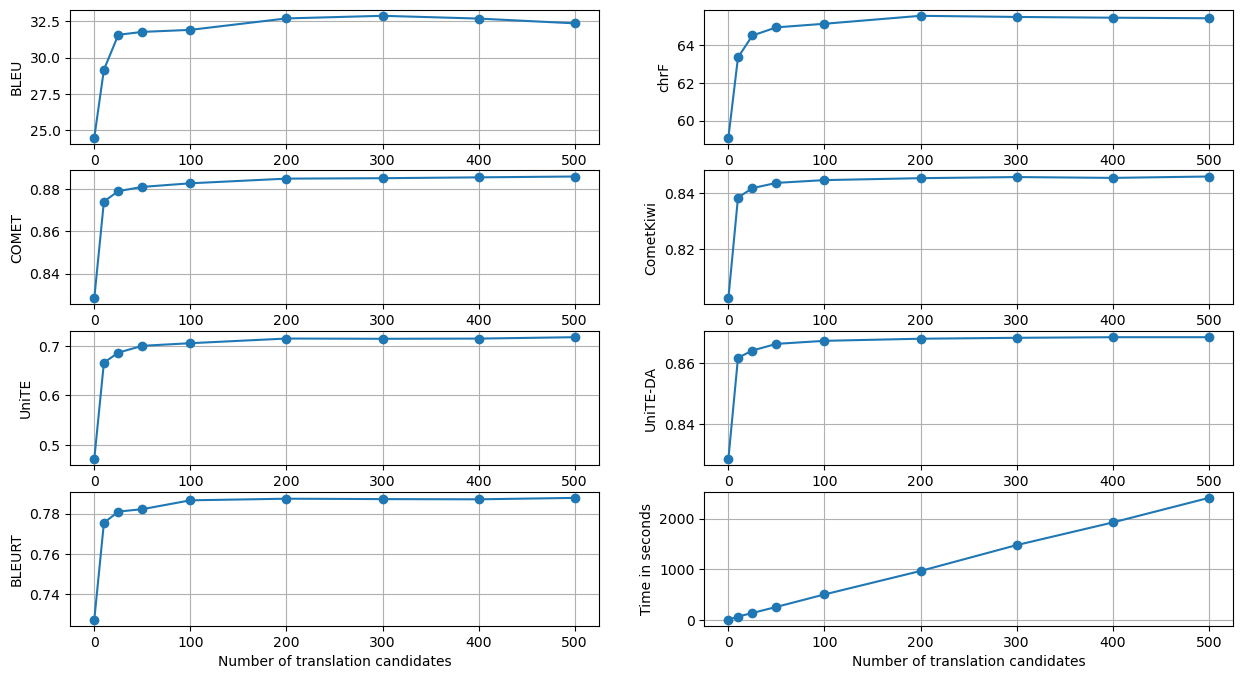}
 \caption{Comparison of top-k performance (temperature 1.0, k=10) with different number of samples of the \mixTune{} English--German model for the \krsTest{} test set. Initial increases in the number of samples for MBR decoding showed very rapid gains, but further increases no longer resulted in such large gains (translation without MBR decoding is represented on the chart as the number of translation candidates equal to 0).}\label{fig:krs-topk10}
\end{figure*}

\begin{figure*}[hbt!]
 \centering
 \includegraphics[scale=0.5]{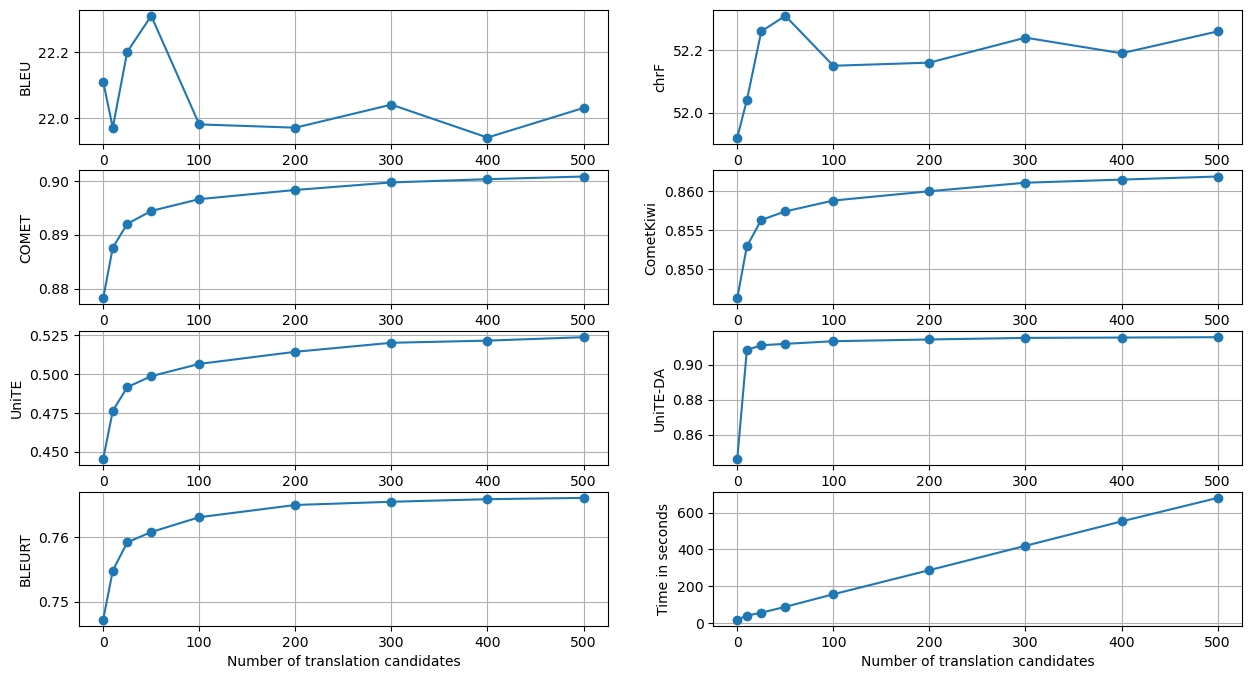}
 \caption{Comparison of beam search performance with different number of samples of the \baseGen{} Czech--Ukrainian model for the \floresTest{} test set. Initial increases in the number of samples for MBR decoding showed very rapid gains, but further increases no longer resulted in such large gains and performance on the n-gram metrics deteriorated (translation without MBR decoding is represented on the chart as the number of translation candidates equal to 0).}\label{fig:cs-uk-bs}
\end{figure*}

\begin{figure*}[hbt!]
 \centering
 \includegraphics[scale=0.5]{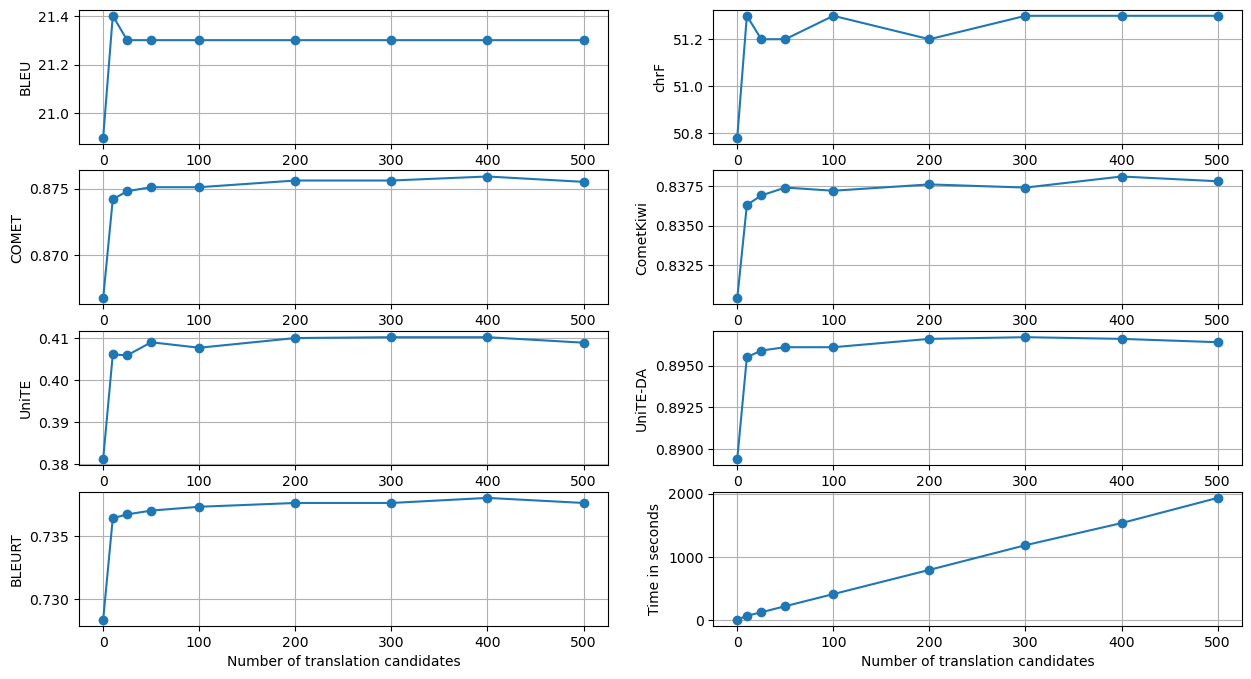}
 \caption{Comparison of top-k performance (temperature 0.1, k=10) with different number of samples of the \baseGen{} Czech--Ukrainian model for the \floresTest{} test set. Initial increases in the number of samples for MBR decoding showed very rapid gains, but further increases no longer resulted in such large gains (translation without MBR decoding is represented on the chart as the number of translation candidates equal to 0).}\label{fig:cs-uk-topk01}
\end{figure*}

\begin{figure*}[hbt!]
 \centering
 \includegraphics[scale=0.5]{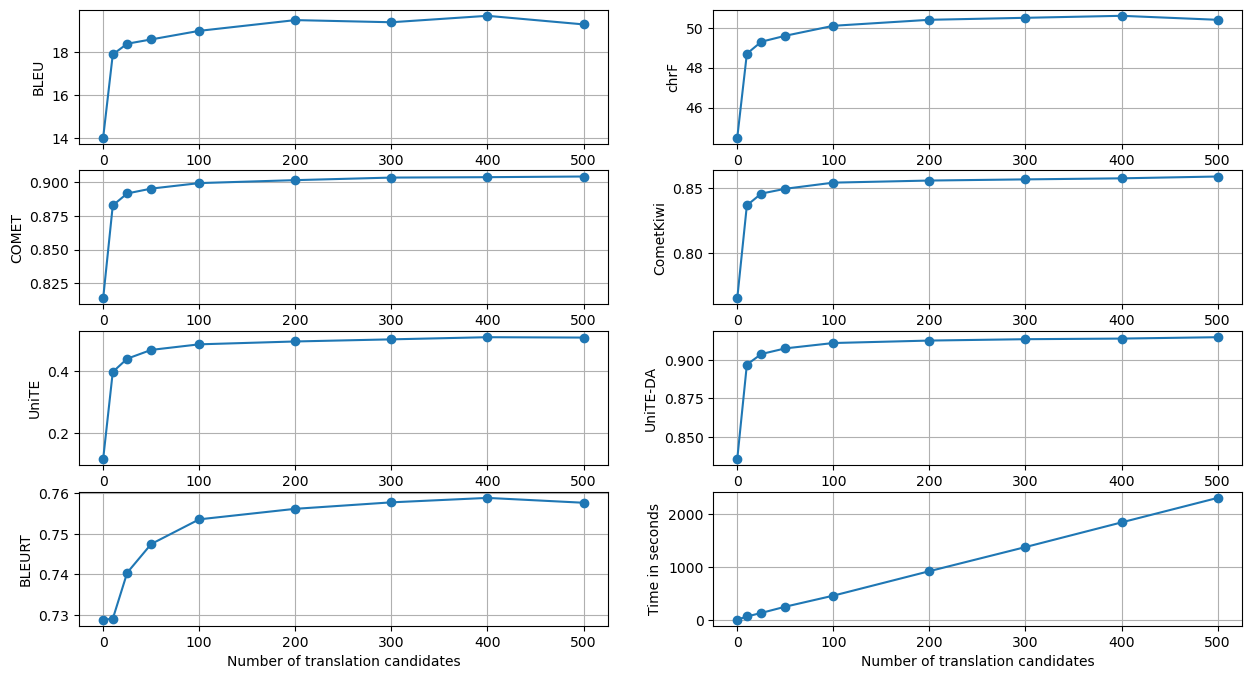}
 \caption{Comparison of top-k performance (temperature 1.0, k=10) with different number of samples of the \baseGen{} Czech--Ukrainian model for the \floresTest{} test set. Initial increases in the number of samples for MBR decoding showed very rapid gains, but further increases no longer resulted in such large gains (translation without MBR decoding is represented on the chart as the number of translation candidates equal to 0).}\label{fig:cs-uk-topk10}
\end{figure*}

\end{document}